\documentclass[10pt,twocolumn,letterpaper]{article}

\usepackage{iccv}
\usepackage{times}
\usepackage{epsfig}
\usepackage{graphicx}
\usepackage{amsmath}
\usepackage{amssymb}

\usepackage[pagebackref=true,breaklinks=true,letterpaper=true,colorlinks,bookmarks=false]{hyperref}

\usepackage{booktabs}       %
\usepackage{amsmath,amssymb}
\usepackage{color}
\usepackage{caption}
\usepackage{subcaption}
\usepackage{graphicx}
\usepackage{xspace}
\usepackage{microtype}

\newcommand\blfootnote[1]{%
  \begingroup
  \renewcommand\thefootnote{}\footnote{#1}%
  \addtocounter{footnote}{-1}%
  \endgroup
}

\providecommand{\shortcite}[1]{\cite{#1}}
\newcommand{\mc}{\mathcal}
\newcommand{\R}{\mathbb{R}}
\newcommand{\bs}{\boldsymbol}
\newcommand{\norm}[1]{\left|\left|#1\right|\right|}

\iccvfinalcopy %

\def\iccvPaperID{5849} %

\ificcvfinal\fi

\begin{document}

\title{Effectively Leveraging Attributes for Visual Similarity}

\author{Samarth Mishra$^{*1}$ \qquad Zhongping Zhang$^{*1}$ \qquad Yuan Shen$^2$ \qquad Ranjitha Kumar$^2$ \\
Venkatesh Saligrama$^1$ \qquad Bryan A.\ Plummer$^1$ \\
$^1$Boston University \qquad $^2$University of Illinois, Urbana-Champaign\\
\small{$^1$\texttt{\{samarthm, zpzhang, srv, bplum\}@bu.edu}} \hspace{0.5cm} \small{$^2$\texttt{\{yshen47, ranjitha\}@illinois.edu}}
}

\maketitle
\ificcvfinal\thispagestyle{empty}\fi

\begin{abstract}
    Measuring similarity between two images often requires performing complex reasoning along different axes (\eg, color, texture, or shape). Insights into what might be important for measuring similarity can can be provided by annotated attributes. Prior work tends to view these annotations as complete, resulting in them using a simplistic approach of predicting attributes on single images, which are, in turn, used to measure similarity. However, it is impractical for a dataset to fully annotate every attribute that may be important. Thus, only representing images based on these incomplete annotations may miss out on key information.
    To address this issue, we propose the Pairwise Attribute-informed similarity Network (PAN), which breaks similarity learning into capturing similarity conditions and relevance scores from a joint representation of two images.  This enables our model to identify that two images contain the same attribute, but can have it deemed irrelevant (\eg, due to fine-grained differences between them) and ignored for measuring similarity between the two images.
    Notably, while prior methods of using attribute annotations are often unable to outperform prior art, PAN obtains a 4-9\% improvement on compatibility prediction between clothing items on Polyvore Outfits, a 5\% gain on few shot classification of images using Caltech-UCSD Birds (CUB), and over 1\% boost to Recall@1 on In-Shop Clothes Retrieval. Implementation available at \href{https://github.com/samarth4149/PAN}{https://github.com/samarth4149/PAN} %
    
\end{abstract}

\blfootnote{*Indicates equal contribution}
\section{Introduction}

\begin{figure}[t!]
    \centering
    \includegraphics[width=0.95\linewidth,trim=0cm 0cm 0cm 0cm,clip]{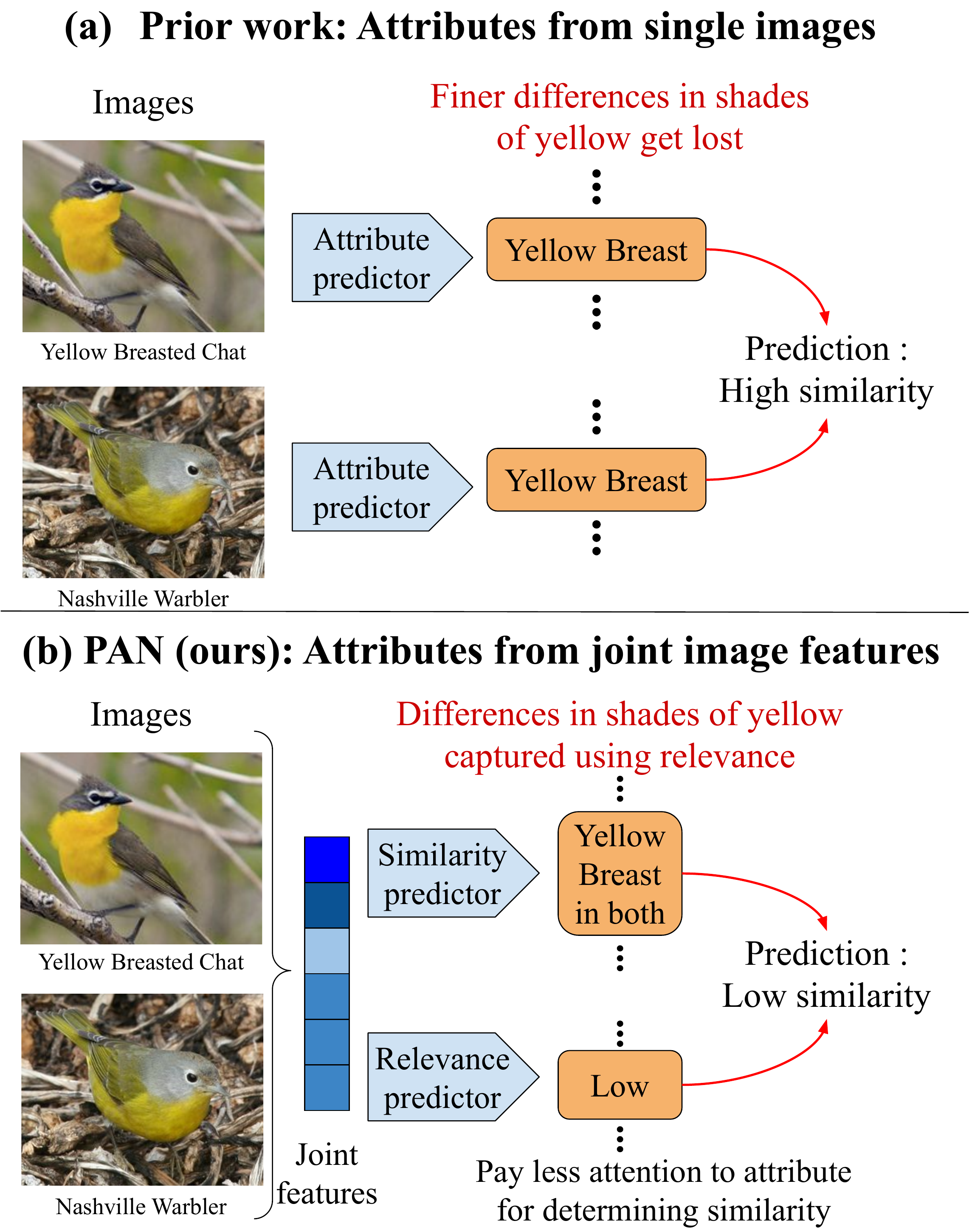}
    \caption{In prior work (\eg~\cite{koh2020concept,plummerPLCLC2017,vasileva2018learning,yang2019interpretable}), shown in (a), attributes used for image similarity are predicted for each image and then are used as input to the image similarity model.  However, this can result in loss of important information about how attributes are expressed (\eg, different shades of the attribute \emph{yellow breast}).  Thus, in our work, shown in (b), we avoid this loss of information by using a joint representation of the two images to compute multiple disentangled similarity scores, each corresponding to an attribute, and relevances of each similarity score in the final similarity prediction. This allows for more fine-grained reasoning about different attribute manifestations, boosting performance.}
    \label{fig:comparison}
\end{figure}

\begin{figure*}[t]
    \centering
    \includegraphics[width=0.8\linewidth]{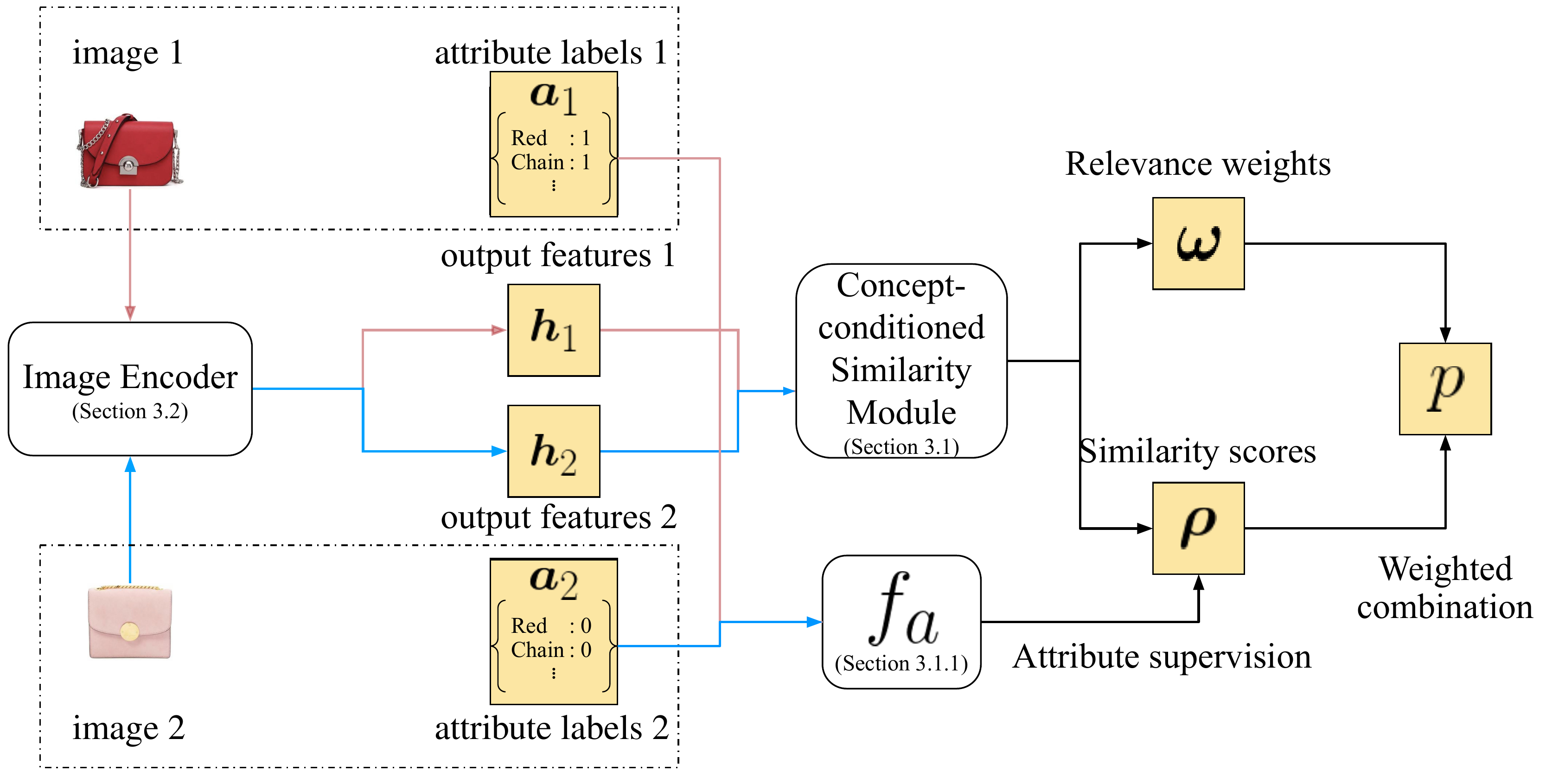}
    \caption{\textbf{PAN overview.}  Given a pair of images, the goal of PAN is produce its similarity score.  We begin by using the image encoder to generate feature vectors for input images. %
    The image features are then fed into the Concept-conditioned Similarity Module (CSM) that uses these features to generate a set of similarity scores with corresponding relevance weights. This enables PAN to identify that two images do contain the same attribute, but that they are not relevant to the similarity score since they are different manifestations of the attribute (see Fig~\ref{fig:comparison} for an example).  The final similarity score $p \in [0, 1]$ is produced using a weighted combination of the similarity conditions and their relevance. Note that the different colored lines (blue, pink) represent information flow pertaining to individual images.  %
    }
    \label{fig:pipeline}
\end{figure*}

Learning similarity metrics between images is a central problem in computer vision with wide-ranging applications such as face recognition \cite{liu2017sphereface,schroff2015facenet}, image retrieval \cite{gordo2016deep,noh2017large,yang2019efficient}, prototype based few shot image classification \cite{koch2015siamese,snell2017prototypical,sung2018learning,vinyals2016matching}, continual learning of image classification \cite{castro2018end,rebuffi2017icarl,stojanov2019incremental}, and fashion compatibility or recommendation  \cite{cucurull2019context,tan2019learning,vasileva2018learning,veit2017conditional,veit2015learning,zhang2018become}. There has been a recent trend of learning these metrics by decomposing the problem into multiple axes of similarity or \emph{similarity conditions}, which has improved performance on a variety of tasks~\cite{HeICDM2016,LIN_2020_CVPR,plummerHint2019,plummer2018conditional,plummerPLCLC2017,tan2019learning,vasileva2018learning,veit2017conditional}. Generally speaking, methods that automatically learn what these conditions represent~\cite{plummer2018conditional,tan2019learning} have reported better performance than those that predefine this knowledge using information like labeled image attributes and item categories~\cite{LIN_2020_CVPR,veit2017conditional,vasileva2018learning,plummerPLCLC2017}. 
We argue this is primarily due to prior work using attributes to predict their presence on single images (\eg~\cite{koh2020concept,yang2019interpretable,vasileva2018learning,plummerPLCLC2017}), and subsequently using these predictions for predicting similarity (Fig \ref{fig:comparison} (a)). This incurs a loss in information about the different manifestations of an attribute, differences that could affect similarity prediction, but may not be distinguishable in attribute annotations. While this could be addressed by collecting a complete set of annotations of every possible attribute and their different manifestations that could affect similarity, such a collection would be expensive.  In addition, it is often impossible to articulate every fine-grained attribute that may affect similarity.

In this paper, we introduce a Pairwise Attribute-informed similarity Network (PAN) that effectively learns to use supervisory information in the form of attribute labels, avoiding information loss, to create a powerful image similarity model that performs well on a range of diverse tasks.   
To illustrate how we do this we refer to the example in Fig \ref{fig:comparison}(b). The figure shows two birds (of different categories) from the Caltech-UCSD Birds (CUB) dataset \cite{wah2011caltech}, where they are both labeled positively for the binary attribute \texttt{has\_breast\_color::yellow}, indicating that they have yellow breasts. Prior work (\eg~\cite{koh2020concept,plummerPLCLC2017,vasileva2018learning,yang2019interpretable}) directly predicts attributes for each image,  which tends to lose information about subtle differences in the manifestations of the attributes, like the shades of the color yellow. 
Our PAN model avoids this issue by first comparing images in a feature space rather than attribute space, as illustrated in Fig~\ref{fig:comparison}(b). Using the joint image features it then predicts both a similarity score and a relevance for different similarity conditions defined by the attributes. Even when the similarity score may coarsely indicate that the two images are similar since they have the same attribute, the model can pick up on finer attribute differences and decide that the mere presence of the same attribute is of low relevance to a positive similarity prediction.
As our experiments will show, this difference can make a dramatic impact on the performance of the learned image similarity model.

One major challenge we face is the considerable difference in how attributes relate to the similarity functions that arise in different tasks.  For example, in few-shot classification, where we use the labeled support images in a nearest-neighbors classifier, the goal of a visual similarity classifier would be to simply measure similarity by matching attributes between the test and support images.  In contrast, for tasks like fashion compatibility, where two images are deemed similar if they complement each other when worn together in an outfit, image pairs with different attributes (\eg, black and orange) can indicate they are highly compatible.  Furthermore, simply modeling which attribute pairs indicate compatibility is insufficient, since two attributes which often result in compatible pairs could be deemed incompatible depending on the other attributes that are present.  For example, \emph{black} and \emph{orange} items are often compatible, except when some other attributes like \emph{red} are also present.  Thus, visual similarity models must learn a far more complex set of relationships between attributes when learning fashion compatibility.  These differences mean that methods that do well on few-shot classification often perform poorly on fashion compatibly and vice-versa. PAN, however, can take this into account via the method we use to convert the incomplete attribute labels for single images into supervisory signals for image pairs and improve performance across diverse tasks.

As we will discuss in Sec \ref{sec:model} (and illustrated in Fig \ref{fig:pipeline}), PAN naturally allows for training and automatically learning similarity conditions in the absence of any additional data like attributes. We also find that PAN can improve performance even in cases where only sparse attribute labels are available.

Summarizing our contributions:
\begin{itemize}
    \item We propose a Pairwise Attribute-informed similarity Network (PAN), which incorporates fine-grained attribute information during training based on a joint representation of two images enabling us to avoid the loss of information suffered by prior work.
    \item While prior methods of incorporating attribute information under-perform prior art, PAN outperforms them on three diverse tasks---by 4-9\% on fashion item compatibility prediction on Polyvore Outfits~\cite{vasileva2018learning}, 5\% on few shot classification on CUB \cite{wah2011caltech} and over 1\% Recall@1 on In-Shop Clothing Retrieval~\cite{liu2016deepfashion}, demonstrating PAN's generality. In comparison to prior approaches of incorporating attribute supervision, PAN is better by a wider margin, \eg, it outperforms them by a sizeable 6-17 \% on Polyvore Outfits.
    \item We propose different methods of using attributes for supervising predictions along similarity conditions, delving into the interpretations of each, providing insights for their applicability in different tasks.  
    \item Our analysis also outlines the contributions of the training procedures, specifically training batch-size. This has commonly been overlooked in prior work, but could have significant impact on final model performance. In doing so, we factor out contribution of the training procedure in demonstrating PAN's benefits. 
\end{itemize}

\section{Related Work}

\noindent\textbf{Visual Similarity Learning.} Learning visual similarity can be used for a wide range of visual tasks. A couple widely used evaluation tasks for similarity learning are face recognition \cite{chopra2005learning,liu2017sphereface,nguyen2010cosine,schroff2015facenet} and image retrieval \cite{noh2017large,gordo2016deep,yang2019efficient}, the latter itself subcategorized multiple ways depending on type of images involved, \eg fashion images, natural images. Image retrieval has a fairly direct application in e-commerce. 
A major portion of this industry consists of websites selling clothing and accessories, giving rise to the challenging task of predicting fashion compatibility \cite{veit2015learning, kang2017visually, vasileva2018learning}, which, as discussed in the introduction, is a form of visual similarity different from the conventional notion. However, we use similarity and fashion compatibility interchangeably as they are addressed in the same manner in our paper.

Some prior visual similarity learning approaches focused on learning a single similarity space~\cite{veit2015learning,kang2017visually,han2017learning}. More recent work \cite{veit2017conditional,plummerHint2019,vasileva2018learning,tan2019learning,LIN_2020_CVPR} found that learning multiple \emph{similarity conditions}, each capturing a different concept, performs better. Concurrent to some of these, the role of using contextual information in image encodings was discussed by Cucurull~\etal~\cite{cucurull2019context}. They used a graph convolutional network (GCN) on images, with similarity links defining the graph structure, to achieve state of the art performance in fashion compatibility prediction. 

Up until now, methods using predefined similarity conditions~\cite{plummerHint2019,veit2017conditional,vasileva2018learning} have underperformed methods that automatically learned these conditions~\cite{tan2019learning,LIN_2020_CVPR}.  Via PAN, we show a method of incorporating additional attribute annotations to supervised similarity conditions while improving final task performance, breaking this trend.
\smallskip

\noindent\textbf{Few Shot Learning.} 
Given the cost associated with acquiring human-annotated labels, learning with few labelled examples is well sought after in computed vision, with a range of prior work exploring possible solutions~\cite{miller2000learning,li2006one,lake2011one,koch2015siamese}. Given a few examples with labeled classes, \cite{vinyals2016matching} classified novel examples using attention weights to compute a probability distribution over known classes. They introduced an episodic training paradigm, 
later also adopted by \cite{ravi2016optimization,snell2017prototypical}. Each episode in an N-way K-shot classification task is a sample of N classes, with K images from each class available to the learning algorithm as ``labeled'' or ``support'' examples for learning few shot classification. Each episode is also accompanied by some query examples which the learner is supposed to predict class labels for. For training, the learning algorithm is provided episodes sampled from a base training dataset, and it is typically evaluated on test episodes sampled from a novel split of the dataset, containing classes different from those appearing in the base split. Both \cite{vinyals2016matching} and \cite{snell2017prototypical} adopted a strategy of minimizing a distance metric in feature space between query images and a prototypical support vector for training. Subsequent work \cite{sung2018learning} improved few shot classification performance by learning a parametric distance measure rather than using a closed-form metric without learnable parameters. 

Another body of work uses meta-learning \cite{finn2017model,Qiao_2019_ICCV,Sun_2019_CVPR} to initialize a classification model's parameters allowing for fast adaptation to a new few-shot task using just a few parameter updates. \cite{chen2019closer} performed a thorough study of several recent few-shot learning methods and proposed a strong baseline classifier using cosine similarity and data augmentation. %
Some recent methods have also employed graph neural networks for few shot classification~\cite{garcia2018fewshot,Kim_2019_CVPR,Schonfeld_2019_CVPR}.  

Like \cite{veit2017conditional, vasileva2018learning, tan2019learning}, PAN lies in the space of decomposing similarity prediction into multiple similarity conditions, but unlike them, PAN can use attribute supervision when available to supervise these conditions and improve performance. PAN is a general similarity learning approach and its benefits with multiple similarity condition spaces and attribute supervision can be seen with different image encoders, be it a simple CNN or a contextual GCN encoder similar to that used by \cite{cucurull2019context}, and on different tasks, like fashion compatibility and few shot classification.

\section{Pairwise Attribute-informed similarity Network (PAN)} \label{sec:model}

 Given two images $\bs x_1$ and $\bs x_2$, the goal of our PAN model is to output a score in $[0, 1]$ representing the probability that the two images are similar.  The primary contribution of our approach lies in its Concept-conditioned Similarity Module (Sec~\ref{subsec:dec-desc}), which takes features  $\bs h_1, \bs h_2 \in \R^d$ representing the input images (computed using an image encoder from Sec~\ref{subsec:enc-desc}), and predicts their similarity $p$ using a weighted combination of predictions along multiple axes of similarity $\bs\rho$ and their relevance $\bs\omega$.  As discussed in the introduction, learning relevance weights over the attribute defined similarity conditions can help us selectively ignore them when said attributes are present, but are not quite relevant in determining similarity between the images. 
 
As our experiments demonstrate, the similarity conditions can be unsupervised as in prior work~\cite{plummer2018conditional,tan2019learning}, but the aim of PAN is to learn to supervise these conditions so they represent a particular concept (\ie, a particular attribute).  These attributes can represent any concept that has been annotated in a dataset.  For example, for CUB~\cite{wah2011caltech} these represent parts of birds (\ie, the type of beak or tail features).  In Polyvore Outfits~\cite{vasileva2018learning} and InShop Retrieval~\cite{liu2016deepfashion} these attributes can contain low-level concepts like colors and textures as well as high-level concepts like ``formal'' and ``fashionable''.  We discuss how we convert the attributes/concepts, which are annotated per image, into labels for image pairs in Sec~\ref{subsec:def-similarity-conds}. Fig \ref{fig:pipeline} provides an overview of our approach.

\subsection{Concept-conditioned Similarity Module (CSM)} \label{subsec:dec-desc}

Given features $\bs{h}_i, \bs{h}_j$ $\in \R^{d}$ of two images, our Concept-conditioned Similarity Module (CSM) generates a set of $M$ similarity scores  $\bs{\rho} = [\rho_1, \ldots, \rho_M] \in \R^M$ and corresponding relevance weights $\bs{\omega} = [\omega_1, \ldots, \omega_M] \in \R^M$ which represent the importance of each similarity condition to the final similarity score:

\begin{align}
    \bs{\rho} &= \sigma \left( \bs{W}_1^\top |\bs{h}_i - \bs{h}_j| + \bs{b}_1 \right) \\ 
    \bs{\omega} &= \text{softmax} \left( \bs{W}_2^\top |\bs{h}_i - \bs{h}_j| + \bs{b}_2 \right)
\end{align}
where $M$ is the number of distinct similarity conditions, $|\cdot|$ represents an element-wise absolute value, and $\sigma(\cdot)$ an element-wise sigmoid function. $\bs{W}_1, \bs{W}_2 \in \R^{d \times M}$, and $\bs{b}_1, \bs{b}_2 \in \R^{M}$ are learnable parameters.  Note that $\bs\rho$ is supervised using attribute labels, but the relevance scores $\bs\omega$ are treated as a latent variable and automatically learned.  
The final similarity score $p \in [0, 1]$ is calculated as the sum of similarity conditions weighed by their relevance, \ie,
\begin{align}
    p = \sum_{m=1}^{M} \rho_m \omega_m = \bs{\rho}^\top \bs{\omega}.
\end{align}

Note that prior work that predicted multiple similarity conditions in both the supervised and unsupervised setting did so based off the features of a single image (\eg,~\cite{LIN_2020_CVPR,plummerHint2019,vasileva2018learning,veit2017conditional}). In contrast, CSM predicts these conditions off a joint representation of both images.  As we show in Appendix \ref{sec:more_questions}, this results in a significant boost to performance when combined with relevance scores.  We believe this is due, in part, to the fact that this joint representation makes it easier to identify differences in attribute manifestations (due to taking the difference of features for the two images). Thus, our approach can more accurately identify when to ignore attribute predictions.

\subsubsection{Defining Similarity Conditions} \label{subsec:def-similarity-conds}

Depending on the availability of labelled attributes for the images, we can choose to supervise similarity conditions to give them semantic meaning. This choice results in two kinds of similarity conditions as described below: 
\smallskip

\noindent\textbf{Unsupervised similarity conditions.}  Similarity conditions are treated as latent variables as done in \cite{tan2019learning}.  The benefit of this approach is that it requires no additional annotations.  Note that all the conditions we predict are based on a joint representation of the two images, unlike in \cite{tan2019learning} where they were computed per-image.
\smallskip

\noindent\textbf{Supervised similarity conditions.} Unsupervised similarity conditions need no attribute annotations. However, we would expect that with some expert knowledge of what might be important in the image, as provided with attributes, we can improve performance using these attributes effectively. Hence, rather than treating each similarity condition as a latent variable, supervised similarity conditions are trained to reflect a specific concept. Since attribute annotations are defined per image, and we predict attributes based off a joint representation, we convert these labels to represent both images, as described next.

Suppose the images have $M$ labeled binary attributes. Each image $i$ is then accompanied with an $M$ dimensional vector $\bs{a}_i \in \{0, 1\}^{M}$. For a pair of images $i$ and $j$, we can use a function $f_{a} : \{0, 1\} \times \{0, 1\} \rightarrow [0, 1]$ to get an $M$ dimensional vector $\bs{a}_{i, j} = f_{a}\left( \bs{a}_i, \bs{a}_j \right)$. Elements of $\bs{a}_{i, j}$ can then be used as labels for supervising the similarity conditions in the model output scores $\bs{\rho}$. Note that if there are missing entries in $\bs{a}_{i, j}$ because of missing attribute labels, these can be handled by zeroing out the loss resulting from them using a binary mask over the indices of $\bs{a}_{i, j}$. %

In Section~\ref{subsec:logical_function}, we experiment with common logical functions as $f_a$, which map to clear semantic meanings. For instance, using $f_a = logical~AND$, with the similarity score $\bs \rho$, the model is asked to predict whether a given attribute appears in both images. Similarly with $OR$, model predicts whether the attribute is in either image, with $XOR$ it predicts whether it is exclusively in one image, and with $XNOR$ it predicts whether the attribute is in both images or in neither. A more detailed discussion of opting for these 4 choices for $f_a$ is in Appendix \ref{sec:more_questions}.  In practice for a given dataset, a logical function can be selected using some prior knowledge about how the attributes relate to the similarity score, or can be selected empirically using held-out data.

\subsection{Types of Image Encoder} \label{subsec:enc-desc}

As mentioned previously, the image encoder generates a lower dimensional feature representation $\bs h$ for an image $\bs x$. We experiment with three different image encoders.

\smallskip

\noindent\textbf{Convolutional Network.}  Unless specified otherwise, we use a simple convolutional neural network (CNN), specifically a ResNet~\cite{he2016deep} to obtain our image feature representation (details in the Appendix \ref{sec:implementation_details}).
\smallskip

\noindent\textbf{Graph Encoder (GE)~\cite{cucurull2019context}.} For some image similarity tasks like fashion compatibility, context can be an important cue in determining how similar two items are. Thus, the second encoder we explore is a graph convolutional network (GCN) that operates on features extracted by a CNN. The GE (composed of the CNN and GCN) takes in as input both images from a dataset and an adjacency matrix over them and simultaneously generates features for all images. This encoder was also used by Cucurull~\etal~\cite{cucurull2019context} and we refer readers to Section 3.1 of their paper for complete details. 
\smallskip

\noindent\textbf{ProxyNCA++~\cite{wern2020proxynca++}.} Many tasks may also find context unhelpful or that GCNs may be too computationally expensive to use~\cite{LIN_2020_CVPR}.  For example, in retrieval tasks a particular emphasis is placed on speed as methods may have to search through millions of images in order to locate the desired item.  As such, for our last encoder we use a state-of-the-art retrieval method when evaluating on the In-Shop Retrieval task~\cite{liu2016deepfashion}. ProxyNCA++ at its core learns a distance metric between images based on learning proxy feature representations for each class. Consequently, it relies on annotated categories for images, and cannot directly be applied for similarity metric learning where no such annotation is available. We refer readers to \cite{wern2020proxynca++} for complete details of this encoder.

\subsection{Model Objective and Training} \label{subsec:model-objective}

The final objective function on a pair of images $\bs x_i$ and $\bs x_j$ is then defined as:
\begin{align}
    \mc{L}(\bs{x}_i, \bs{x}_j, e_{i, j}, \bs{a}_{i, j}) = \mc{L}^{BCE} (e_{i, j}, p) + \lambda \mc{L}_{el}^{BCE} (\bs{a}_{i, j}, \bs{\rho}),
    \label{eq:total_loss}
\end{align}
where $\lambda$ is a tunable hyperparameter, $e_{i, j} \in \{ 0, 1 \}$ is the ground truth similarity label between images $\bs x_i$ and $\bs x_j$, $\mc{L}^{BCE}$ is the binary cross-entropy loss and $\mc{L}_{el}^{BCE}$ is the mean element-wise binary cross-entropy. Note that when there are no supervision attributes, the second term in Equation \ref{eq:total_loss} is $0$. For training, an equal number of positive and negative pairs are sampled randomly from the training split and the model is trained to predict similarity between them. Details regarding the exact process for each encoder are in Appendic \ref{sec:implementation_details}.

\section{Datasets and Tasks}

\noindent\textbf{Polyvore Outfits~ \cite{vasileva2018learning}} contains 53K outfits (sets of fashion items) for training, 5K for validation and 10K for testing. It also provides fine-grained category information and text descriptions of items.  We use the 205 sparsely annotated attributes from \cite{plummerSimilarity} as labels for supervising similarity conditions. %
Evaluation involves two tasks.  First, in the fill-in-the-blank (FITB) outfit completion task a model is given a partial outfit and has to select from four possible answers what item would best complete it.  Performance is measured by how often the answer was correct. Second, in outfit compatibility a model is asked to discriminate between good and bad outfits. Performance is measured using area under a receiver operating characteristic curve (AUC). Following~\cite{vasileva2018learning,cucurull2019context}, outfit compatibility scores are computed by averaging the similarity prediction over all pairs of items in the outfit. There are 10K FITB questions and 10K each positive and negative samples for outfit compatibility (20K total) in the test split.  

Since current methods get almost perfect performance on the original outfit compatibility task, we created a more challenging testing set of the same size by modifying the procedure outlined in \cite{vasileva2018learning}, that we refer to as the \emph{resampled} set. For outfit compatibility we collected new negative outfits by replacing only part of a ground truth outfit, unlike the original split which replaced all items.  We randomly selected the number of items to replace, and each item is replaced with another of the same type in the same split (\ie, a top could only be replaced with another top).   Similarly, we made a more challenging FITB task, where a model must select between 10 candidate answers (the original test had 4).  As with the original sampling, we ensured any replaced items and candidate answers were of the same type.
\smallskip

\noindent\textbf{CUB-200-2011 \cite{wah2011caltech}} consists of 200 classes and a total of 11,768 images of birds. We use the split provided by \cite{chen2019closer} for our experiments, which contains 100 base classes, 50 validation and 50 novel classes. The CUB dataset also has 312 fine-grained binary attributes labeled for each image, with an accompanying score on a 4 point scale indicating the confidence of the assigned label. We drop all attribute labels that have a confidence score less than or equal to 2.

We use the 5-way 5-shot classification task for evaluation. Reported accuracies are averaged over 3 training runs from different random initializations accompanied by 95\% confidence intervals. A test \emph{episode} consists of a random sample of 5 classes and 5 support images from the 50 classes in the novel split of the dataset. 16 query images, distinct from support images, are also sampled for each of these 5 classes. The accuracy for an episode is the 5-way accuracy of a classifier over the 16 x 5 = 80 query images. A few shot learning model is evaluated using its average classification accuracy over 600 randomly generated test episodes. 
\smallskip

\noindent\textbf{In-Shop Retrieval \cite{liu2016deepfashion}} contains 52,712 images of clothes from 11,967 classes. There are 14,218 query images and 12,612 gallery images for testing. Given a query image, the task is to retrieve an image of the same item from the gallery set. Note that the query and gallery sets do not overlap with the training set. There are 463 attributes of clothes in total, we use these attribute labels for our PAN-Supervised model. Methods are ranked based on Recall@1.

\section{Results} \label{subsec:results}

\subsection{Comparison with prior work}
Table \ref{tab:compare_prior_work_po}, Table \ref{tab:compare_prior_work_cub}, and Table \ref{tab: inshop_results} compare the best settings (encoder, number of unsupervised similarity conditions, \etc) used by our model %
to representative state-of-the-art results reported in prior work on Polyvore Outfits, CUB, and In-Shop Retrieval, respectively. As shown in Table~\ref{tab:compare_prior_work_po} we obtain a 4\% better FITB accuracy and 9\% AUC boost over the state-of-the-art on the fashion compatibility task using our more challenging resampled test set for both tasks, while also increasing FITB accuracy by 8\% on the original split.  Similarly, in Table~\ref{tab:compare_prior_work_cub} and Table \ref{tab: inshop_results} we observe a 5\% and 1\% performance improvement over the state-of-the-art on fine-grained few shot classification and In-Shop Retrieval, respectively. Improvement over the diverse set of tasks demonstrates PAN's ability to generalize.  Our PAN model can also be useful when no supervision is provided, as our PAN-Unsupervised model obtains a 3-4\% gain over prior work on Polyvore Outfits and CUB, while also boosting performance on In-Shop Retrieval.  Note that fashion compatibility benefited from using a graph image encoder (GE), while few-shot classification reported best performance with a CNN encoder, which we shall discuss further in Section~\ref{subsec:context}.

In addition to comparison with prior work, Table \ref{tab:compare_prior_work_po}, Table \ref{tab:compare_prior_work_cub}, and Table \ref{tab: inshop_results} also provide two alternative methods of using attributes in an image similarity model.  In ``X + Attr.\ Multitask'' we use a hard parameter sharing multitask approach~\cite{caruanaMULTI-task}, where we share an image encoder, but have separate output heads for each of the two tasks (one of them being attribute classification, the other similarity link prediction). In ``Attr.\ Similarity'' we use a standard framework where we predict attributes for each image and then learn a classifier implemented as a fully connected layer that takes both attributes as input and predicts similarity (the general framework used by~\cite{farhadi2009describing}).  Notably, both baseline methods that use attributes only improve performance on few-shot classification, but either make no difference, or are even harmful to performance on the other two datasets (\eg, Attr. Similarity \emph{reduces} FITB performance by 5-8\% compared with the CGAE baseline).  In contrast, our PAN-supervised model outperforms all other methods, including on the fashion compatibility task where we report a staggering 6-17\% boost over the attribute baselines on the resampled test set.
\smallskip

\begin{table}[t]
    \centering
    \scalebox{0.9}{
        \begin{tabular}{|r|l|cc|cc|}
        \hline
        & & \multicolumn{2}{c|}{ Original} & \multicolumn{2}{c|}{Resampled}\\
        \cline{2-6}
        &\textbf{Method} & \textbf{FITB} & \textbf{AUC} & \textbf{FITB} & \textbf{AUC}\\
        \hline
            \textbf{(a)} 
            & TAN~\shortcite{vasileva2018learning} & 57.6 & 0.88 & 38.1 & 0.66\\
            & SCE-Net~\shortcite{tan2019learning} & 61.6 & 0.91 & 43.4 & 0.68\\
            & CSA-Net~\shortcite{LIN_2020_CVPR} & 63.7 & 0.91 & -- & --\\
            & CGAE~\shortcite{cucurull2019context} & 74.1 & \textbf{0.99} & 60.8 & 0.67\\
        \hline
            \textbf{(b)}
            & X + Attr. Multitask-GE & 73.8 & \textbf{0.99} & 57.6 & 0.65\\
            & Attr. Similarity-GE & 69.5 & 0.98 & 52.9 & 0.65\\
            & PAN-Unsupervised-GE & 78.4 & \textbf{0.99} & 64.1 & 0.70\\
            & PAN-Supervised-GE & \textbf{82.3} & \textbf{0.99} & \textbf{69.7} & \textbf{0.71}\\
        \hline
        \end{tabular}
    } %
    \caption{Comparison of PAN on fashion compatibility on Polyvore Outfits to (a) results reported in prior work or reproduced with the author's code and (b) other PAN and attribute supervision approaches.}
    \label{tab:compare_prior_work_po}
\end{table}

\begin{table}[t]
    \centering
    \begin{tabular}{|r|l|c|}
        \hline
        & \textbf{Method} & \textbf{Accuracy} \\
        \hline
        \textbf{(a)}
        & Baseline++ \shortcite{chen2019closer} & 83.58 \\
        & ProtoNet \shortcite{snell2017prototypical} & 87.42 \\
        & TriNet~\shortcite{semanticAugmentation} & 84.10\\
        & TEAM~\shortcite{Qiao_2019_ICCV} & 87.17 \\
        & CGAE~\shortcite{cucurull2019context} & 88.00 $\pm$ 1.13\\
        \hline
        \textbf{(b)}
        & X + Attr. Multitask-GE & 89.29 $\pm$ 0.57\\
        & Attr. Similarity & 92.21 $\pm$ 0.21\\
        & PAN-Unsupervised & 92.60 $\pm$ 0.10\\
        & PAN-Supervised & \textbf{92.77} $\pm$ 0.30  \\
        \hline
    \end{tabular}
    \caption{Comparison of PAN on 5-way 5-shot classification on CUB-200-2011 to (a) results reported in prior work or reproduced with the author's code and (b) other PAN and attribute supervision approaches. Intervals provided are 95\% confidence intervals over 3 different runs with different random model initializations}
    \label{tab:compare_prior_work_cub}
\end{table}

\begin{table}[t]
    \centering
    \setlength{\tabcolsep}{2pt}
    \begin{tabular}{|r|l|c|}
        \hline
        & \textbf{Method} & \textbf{Recall@1} \\
        \hline
        \textbf{(a)}
        & MS \cite{wang2019multi} & 89.7 \\
        & NormSoftMax\cite{zhai2018classification} & 89.4 \\
        & HORDE \cite{jacob2019metric} & 90.4  \\
        & Cont. w/M~\cite{wang2020xbm} & 91.3 \\
        & ProxyNCA++\cite{wern2020proxynca++} & 90.9 \\
        \hline
        \textbf{(b)}
        & ProxyNCA++ \& Attr. Multitask & 90.8 \\
        & ProxyNCA++ \& Attr. Similarity & 86.4 \\
        & ProxyNCA++ \& PAN-Unsupervised & 91.4 \\
        & ProxyNCA++ \& PAN-Supervised & \textbf{92.1} \\
        \hline
    \end{tabular}
    \caption{Comparison of PAN on In-Shop Clothing Retrieval to (a) results reported in prior work and (b) other PAN and attribute supervision approaches.}
    \label{tab: inshop_results}
\end{table}

\begin{figure}[t]
    \centering
    \begin{subfigure}[t]{0.5\textwidth}
    \centering
    \includegraphics[width=\linewidth]{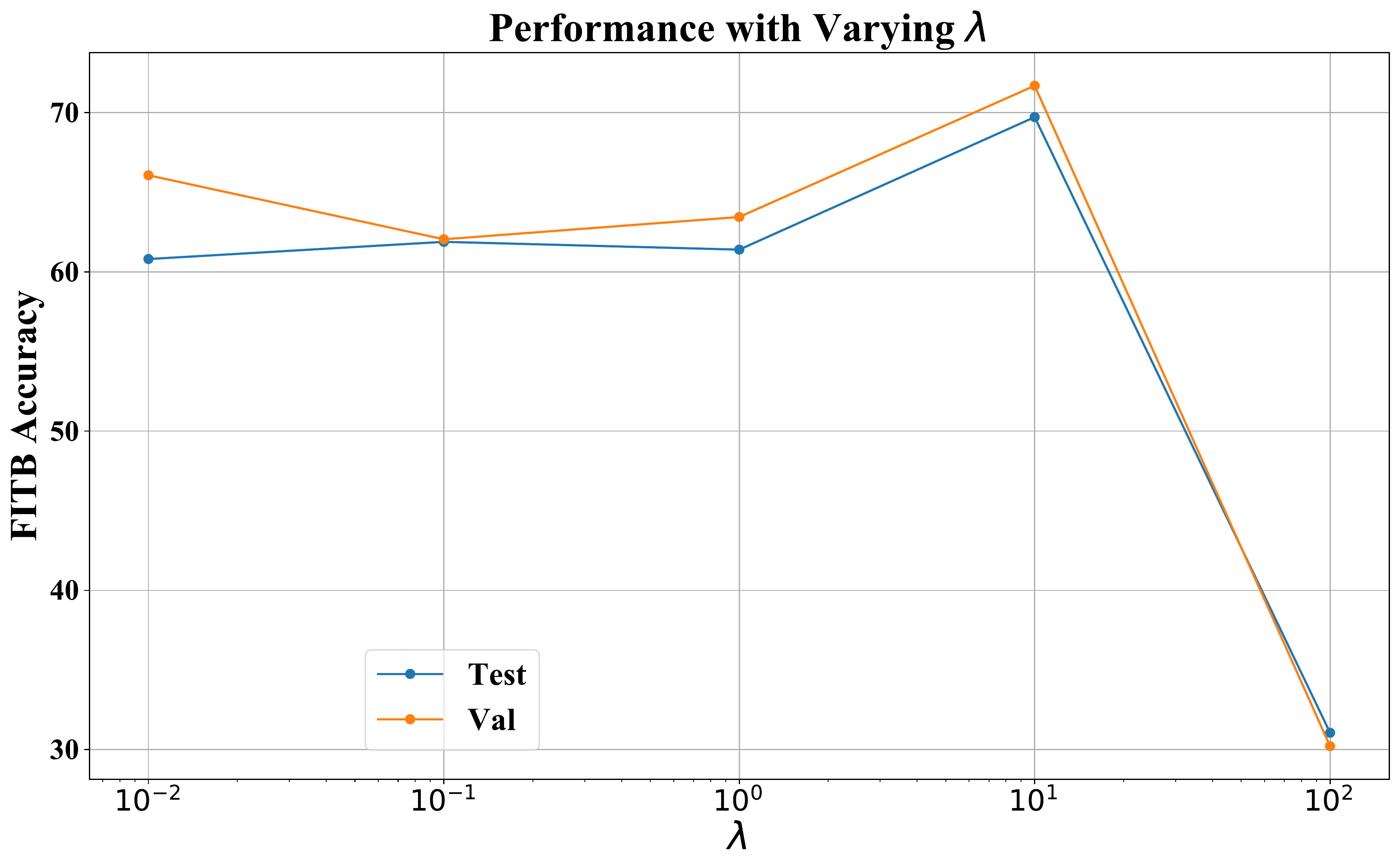}
    \caption{FITB accuracy of PAN-Supervised-GE on Polyvore Outfits.}
    \end{subfigure}
    \begin{subfigure}[t]{0.5\textwidth}
    \centering
    \includegraphics[width=\linewidth]{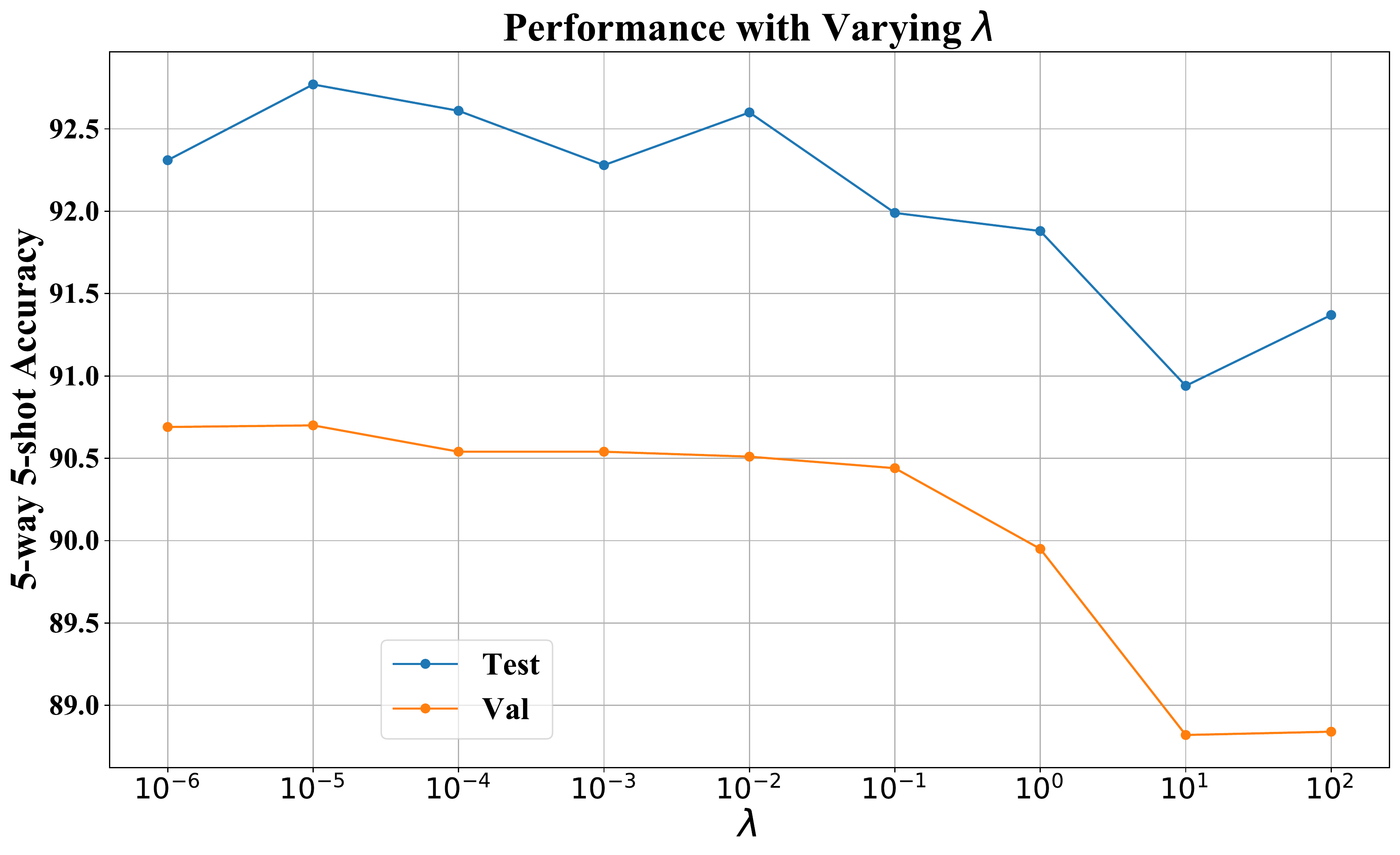}
    \caption{5-way 5-shot classification accuracy of PAN-Supervised on CUB}
    \end{subfigure}
    \caption{Sensitivity to $\lambda$ from Eq.~(\ref{eq:total_loss}). }
    \label{fig:plot-lam}
    \vspace{-5mm}
\end{figure}

\begin{table*}[t!]
    \centering
    \begin{tabular}{|l|c|ccc|cc|}
    \hline
    & Single Batch & \multicolumn{3}{c|}{Polyvore Outfits-Resampled} & \multicolumn{2}{c|}{CUB-200-2011}\\
    \cline{3-7}
    Method & Training & $~~~\bs{M}~~~$ & FITB & AUC & $\bs{M}$ & Accuracy\\
    \hline
    Siamese Network & -- & -- & 36.4 & 0.63 & -- & 76.87 $\pm$ 0.72 \\
    X + Attr. Multitask & -- & -- & 37.2 & 0.63 & -- & 81.96 $\pm$ 1.01 \\
    \hline
    Siamese Network & \checkmark & -- & 44.7 & 0.69 & --  & 89.01 $\pm$ 0.25\\
    X + Attr. Multitask & \checkmark & -- & 46.1 & 0.69 & -- & 75.82 $\pm$ 0.42 \\
    Attr. Similarity & \checkmark & -- & 31.1 & 0.63 & -- & 92.21 $\pm$ 0.21 \\
    PAN-Unsupervised & \checkmark & 50 & 27.3 & 0.62 & 200 & 92.60 $\pm$ 0.10 \\
    PAN-Supervised & \checkmark & 206 & 28.2 & 0.62 & 312 & \textbf{92.77} $\pm$ 0.30\\
    \hline
    X + Attr. Multitask-GE & \checkmark & -- & 57.6 & 0.65 & -- & 89.29 $\pm$ 0.57 \\
    Attr. Similarity-GE & \checkmark & -- & 52.9 & 0.65 & -- & 87.02 $\pm$ 1.27 \\
    PAN-Unsupervised-GE & \checkmark & 50 & 64.1 & 0.70 & 200 & 89.55 $\pm$ 0.48 \\
    PAN-Supervised-GE & \checkmark & 206 & \textbf{69.7} & \textbf{0.71} & 312 & 90.16 $\pm$ 0.51 \\
    \hline
    \end{tabular}
    \caption{Effect of batch size and image encoder on performance on the fashion compatibility and few-shot tasks. GE refers to the graph image encoder. Refer to Section \ref{subsec:context} for discussion.}
    \label{tab:analysis}
\end{table*}

\begin{table*}[t]
    \centering
    \begin{tabular}{|l|cc|cc|c|c|}
    \hline
    & \multicolumn{4}{c|}{Polyvore Outfits} & CUB-200-2011 & In-Shop\\
    \cline{2-7}
    & \multicolumn{2}{c|}{Original} & \multicolumn{2}{c|}{Resampled} & &\\
    \cline{2-5}
    Attribute supervision label ($f_{a}$) &  FITB & AUC & FITB & AUC & Accuracy & R@1\\
     \hline
     \hline
    Present in either ($OR$) & \textbf{82.4} & \textbf{0.99} & \textbf{69.7} & \textbf{0.71}  & \textbf{92.77} $\pm$ 0.27 & 91.5\\ 
    Present in both ($AND$) & 76.3 & 0.98 & 62.4 & 0.62 & 92.61 $\pm$ 0.36 & 91.6\\ 
    Present in both or neither ($XNOR$) & 76.1 & \textbf{0.99} & 60.8 & 0.66 & 92.60 $\pm$ 0.41 & \textbf{92.1}\\
    Present exclusively in one ($XOR$) & 69.0 & 0.98 & 51.8 & 0.65 & 92.61 $\pm$ 0.20 & 91.6\\
    $AND$ concat $XOR~^{*}$ & 78.9 & \textbf{0.99} & 64.7 & \textbf{0.71} & 92.39 $\pm$ 0.13 & 91.9\\
    \hline
    \end{tabular}
    \caption{Effect of different kinds of attribute supervision resulting from different functions $f_a$. $^{*}$includes twice the number of similarity conditions as others}
    \vspace{-5mm}
    \label{tab:att_comb_type}
\end{table*}

\subsection{Choice of image encoder and batch size}
\label{subsec:context}

Prior work has been inconsistent in its training methods and controlling for hyperparameters like batch size, which can significantly affect performance.  Table~\ref{tab:analysis} remedies this by comparing training with the whole training split vs. using minibatches. It also compares the effect of using a graph encoder (GE) instead of a simple CNN. We see comparing the numbers of row 3 of Table \ref{tab:analysis} to those of prior methods in Tables \ref{tab:compare_prior_work_po} and \ref{tab:compare_prior_work_cub} that training with the whole training set can significantly improve performance, making even a simple Siamese network trained using a triplet loss outperform most recent methods on both tasks. We note here that in our experiments with single batch training, we use a pre-trained CNN to extract image features, and do not finetune it.

It is also notable that using GE performs worse than the simpler Siamese Network baseline on CUB, which we believe is due to differences between tasks. Specifically, in fashion compatibility links exist between items that may be very different from each other. Thus, the additional context provided through the GE may be more important than in CUB, which has less variation between linked items since they all contain the same bird.

\subsection{Choice of attribute combination $f_a$}
\label{subsec:logical_function}

As discussed in Section \ref{subsec:def-similarity-conds}, choosing $f_a$ as one of common logical functions can lead to different interpretations of the attribute supervision provided.
Table~\ref{tab:att_comb_type} compares these functions that convert image attribute labels to pairwise labels for use in our PAN model.

At first glance, predicting $1$ when either both images have an attribute or neither of them does (\ie using $XNOR$) seems like an intuitive choice. This would indicate how many attributes match between two images. %
However, some functions like fashion compatibility, where images may match \emph{because} they contain different attributes, $XNOR$ would not be appropriate since any non-shared attributes would be ignored (\ie have a supervision label $0$). $AND$, which encourages models to predict $1$ only when both images have the attribute has same issue. However $OR$ would not face the issue since it can be $1$ when either of the images has an attribute, allowing the model to use its relevance weights to decide whether a combination of attributes is relevant for similarity. In Table \ref{tab:att_comb_type}, we see $OR$ perform the best on fashion compatibility on Polyvore Outfits.

Notably, we also see that $OR$ is still competitive with $XNOR$ on the CUB dataset, where the goal is to determine similarity in the more conventional sense, \ie, similar images should have matching attributes. This task seems like a good fit for $XNOR$. 
However, on CUB, many of the attributes are mutually exclusive (\eg, a bird either has a ``red beak'' or a ``black beak'', but not both). 
If a model can reason jointly about dissimilar attributes inferring that they should not coexist, it can correctly function on this task. Relevance weights in our model allow for this joint reasoning over different attributes.
Thus, training the model to predict $1$ when either image has an attribute ($OR$ supervision), can also perform well on this task, and from the empirical results in Table \ref{tab:att_comb_type}, we see that it does. 

Providing both the $AND$ and $XOR$ outputs (as a concatenation) seems lucrative since it seems more informative than $OR$, but we found that the model uses its additional capacity to overfit to training data. This is also challenging to use since attribute predictions are noisy (see Appendix \ref{sec:attr_recog} for attribute recognition performance).

The In-Shop retrieval task involves fetching matches from a gallery of different views of an object. $XNOR$ is ideal for matching different views in this case, since matching attributes can be directly translated to the inference that two views belong to the same object, therefore, are similar.

\subsection{PAN sensitivity to $\lambda$}
\label{subsec:lambda}
Figure \ref{fig:plot-lam} shows PAN's sensitivity  to different weights ($\lambda$) of the attribute supervision loss. Performance is plotted for both the testing and validation sets. On Polyvore Outfits, we see that the PAN-Supervised-GE model performs well when attribute supervision weight is relatively high (best accuracy at $\lambda=10$). Model performance decreases on either side of this, with a heavier decrease when $\lambda$ is increased significantly (to $100$). On the CUB dataset, we see a somewhat different behavior where the best model performance is achieved at $\lambda=10^{-5}$, which is much lower, indicating relatively lower attribute supervision is optimal for the task.  Note that on the In-Shop task we set $\lambda=1$ and did not tune it on that task, demonstrating that PAN can be readily adapted to improve performance on other tasks/models.

\section{Conclusion}
We presented PAN, a method of incorporating additional attribute annotations in image datasets to learn a better similarity predictor. We saw that PAN's method of decomposing similarity prediction into multiple conditions is general, functions with a range of different image encoders and is flexible in using attribute annotation, possibly sparse, when available. PAN outperformed state of the art on a diverse set of three tasks---by 4-9\% on fashion item compatibility prediction on Polyvore Outfits, 5\% on few shot classification on CUB and over 1\% Recall@1 on In-Shop Clothing Retrieval---contrary to prior approaches of using attribute supervision, which were unable to outperform methods that automatically learned concepts in different similarity conditions. In showing these benefits of PAN we factored out contributions from training parameters like batch-sizes, hopefully informing future work with our analysis.
\smallskip

\noindent\textbf{Acknowledgements:} This work was funded in part by DARPA and by the Hariri Data Science Faculty Fellowship.

{\small
\bibliographystyle{ieee_fullname}
\bibliography{references}
}

\section*{Appendices}
\appendix

\section{Similarity as edge prediction}

To accommodate training with the graph encoder (GE), we formulate learning similarity as an edge prediction task on graph with nodes as images as done in \cite{cucurull2019context}.  Each node in the graph represents an image and the edges represent ground truth similarity information. Edges are stored as an adjacency matrix $\bs{A} \in \R^{N \times N}$ where $\bs{A}_{i, j} = 1$ if there exists an edge between node $i$ and node $j$, \ie, if the images corresponding to the nodes are labelled to be similar in the dataset. $\bs{A}_{i, j} = 0$ otherwise. Note that this is a general formulation and encoders other than the GE can be trained this way. The graph defined just does not play any role in the model's outputs in that case. 

\section{Similarity prediction for end tasks}
\noindent\textbf{Fashion Compatibility Prediction.}
The goal of this task is to predict whether a given set of items is compatible. We compute the compatibility score of an outfit (group of items) by averaging the likelihood of edge existence over all pairs of items in the outfit. Area under a receiver operating characteristic curve (AUC) is used here to evaluate the performance on this task. For this task, at inference time, a model with GE uses no ground truth context information in the form of edges since none is available.
\smallskip

\noindent\textbf{Fill-in-the-blank (FITB).}
The FITB task is to select the best compatible item given a partial outfit and a set of candidate items. Concretely, a question consists of $n$ items ${q_1, q_2, ..., q_n}$. Each question has $m$ choices ${o_1, o_2 ..., o_m}$. Our models compute the compatibility score $s$ between all item pairs and $s_{ij}$ represents the score between $q_i$ and $o_j$. The score of $o_j$ is calculated as $\sum^{n}_{i=0} s_{ij}$. The item that obtains the highest score is chosen as our final candidate.  Performance is measured in terms of answer accuracy. When a GE is used, edges are added to represent compatibility between the items in the question, \ie, edges are added between each $q_i$ and $q_j$ for $i, j \in [n]$, at inference time.
\smallskip

\noindent\textbf{Few shot classification.}
Each 5-way 5-shot classification episode has 5 support examples or 5 ``shots'' for each of the 5 classes and 16 query examples from each class. The task is to classify query examples into one of the 5 classes. For a given query example, the model predicts the probability of existence of an edge between it and the support examples for each of the classes. The average edge probability over support examples for a given class is treated as a score of belongingness to that class. The model then predicts the class of the given example as the one with the highest score. The accuracy of the model for an episode is its accuracy in classifying the 16 x 5 = 80 query examples.

\section{Implementation details} \label{sec:implementation_details}
In this section we describe the training details of our models and different baselines referred to in Section 4.1 in the main paper.

\subsection{Pairwise Attribute-informed similarity Network (PAN)}
Our PAN model uses image features encoded by a pre-trained CNN. We use a Resnet-18 \cite{he2016deep} feature extractor for few shot classification on CUB and a Resnet-50 for fashion compatibility prediction on the Polyvore Outfits dataset. The Resnet-50 used is pretrained on Imagenet \cite{deng2009imagenet}. However, we use the feature extractor from a Siamese Network (details below in section \ref{subsec:appendix-siamese}) for CUB since the novel split of CUB shares some images with the Imagenet dataset. The size of the image input to these feature extractors is 224 x 224, and the features output from Resnet-18 are 512 dimensional while those output from Resnet-50 are 2048 dimensional.

For models evaluated on CUB, the graph encoder contains a 3 layer graph convolutional network (GCN) where the features output at each layer are 350 dimensional. The same for Polyvore Outfits is a 2-layer GCN, the features being 200 dimensional at the output of each layer. 

In the training of models using the graph encoder, we use dropout with drop probability of 0.5 at each GCN layer. As an additional method of regularization, in each epoch of training, each edge in the adjacency matrix $\bs{A}$ is dropped with a probability $0.15$.

For training the PAN models, we use an Adam~\cite{kingmaAdam2015} optimizer with learning rate 0.001. The models for few shot classification on CUB are trained for 1000 epochs and the ones for fashion compatibility prediction on Polyvore Outfits for 4000 epochs, validation being done after each epoch of training.

\subsection{Siamese Network for few shot classification} \label{subsec:appendix-siamese}
We train a Siamese Network~\cite{bromley1994signature, koch2015siamese} for few shot classification on CUB, primarily as a CNN feature extractor for the PAN models.  Training examples are obtained by sampling triplets of images ($\textbf{x}$, $\textbf{y}, \textbf{z}$), where $\textbf{x}$ and $\textbf{y}$ belong to the same class and $\textbf{z}$ belongs to a different class. The network is trained to minimize the following triplet loss
\begin{align}
    \mc{L} = \max(\norm{f(\textbf{x}) - f(\textbf{y})}_2 - \norm{f(\textbf{x}) - f(\textbf{z})}_2 + \alpha, 0)
\end{align}
where $f$ is the ResNet-18 feature extractor, $\norm{\cdot}_2$ is the $\ell_2$ norm, and $\alpha$ is a margin parameter.

We use the same splits for CUB as \cite{chen2019closer}, 100 classes in the  base split and 50 each in the novel and val splits. In one epoch of training, the network sees all possible positive pairs of images in the training dataset (base split of CUB), with negative examples sampled randomly from one of the remaining classes. We trained the network for 200 epochs, validating every 2 epochs using average accuracy on 100 few shot classification episodes drawn from the val split. We used Adam \cite{kingmaAdam2015} optimizer with a learning rate of $0.001$ and the margin parameter $\alpha$ was chosen to be $0.2$. A mini-batch size of 96 triplets was used for training. We used random resizing and cropping, color jittering and random horizontal flips as data augmentation for training.

\subsection{Single batch training}
When training with a single batch (\ie, the entire dataset is used for training at once), fine-tuning the CNN with limited GPU memory is not possible.  Thus, to minimize the GPU memory required for each image, we use a pretrained CNN to get fixed-length feature representations for images.  For methods like a Siamese Network and ``X + Attr Multitask'' we train a classifier implemented as a fully connected layer to predict links between images, using a single batch consisting of pre-extracted features from the entire training set. These features are extracted from a CNN trained using mini-batches either as a Siamese Network or with an additional attribute prediction head, as is done for ``X + Attr Multitask''.

\section{PAN --- Behavior with different number of similarity conditions} \label{subsec:nodes}

\begin{figure*}[t!]
    \centering
    \begin{subfigure}[t]{.39\textwidth}
    \includegraphics[width=\linewidth]{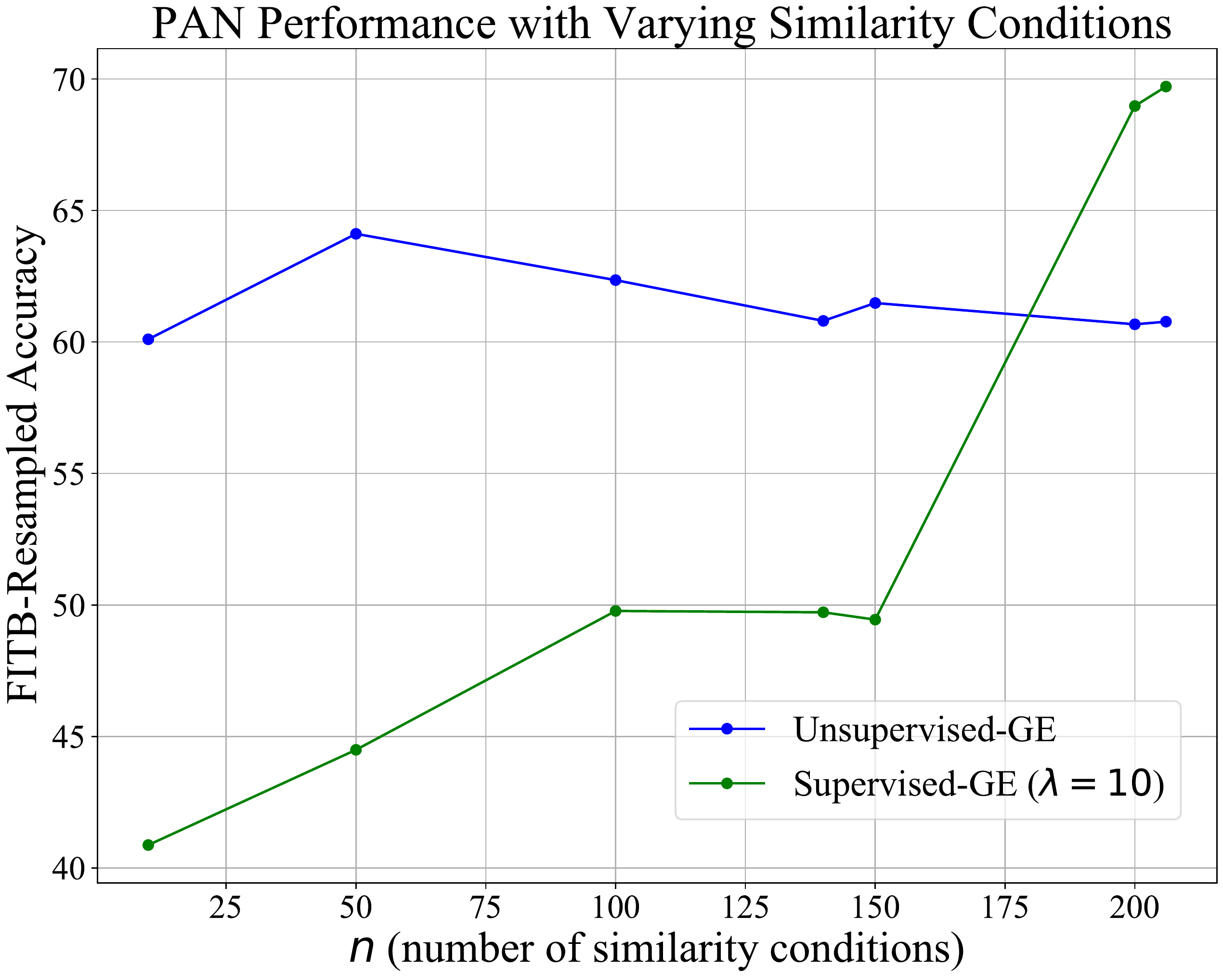}
    \label{fig:plot-nodes-sub1}
    \caption{FITB accuracy on Polyvore Outfits}
    \end{subfigure}
    \begin{subfigure}[t]{.58\textwidth}
    \includegraphics[width=\linewidth]{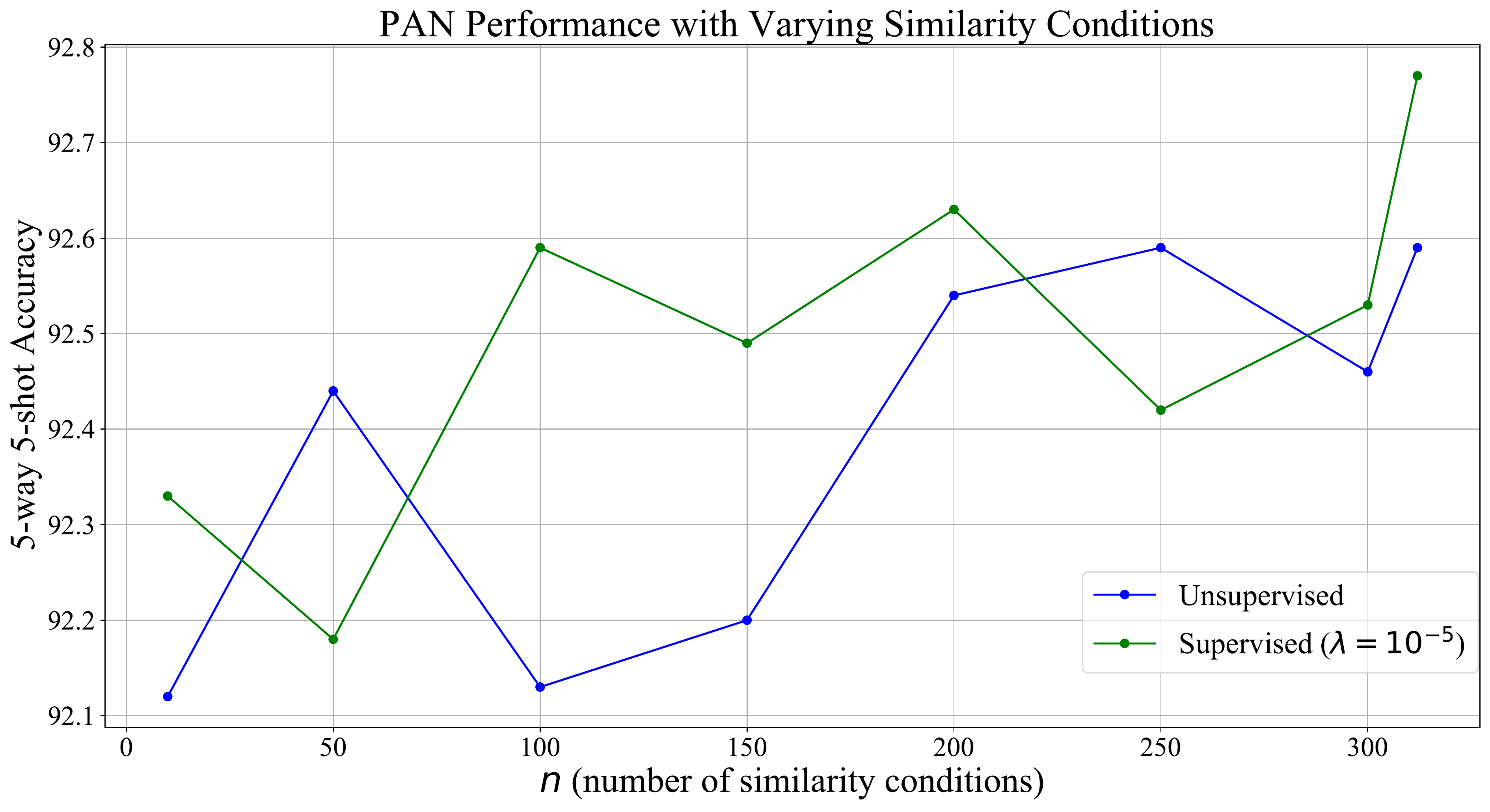}
    \label{fig:plot-nodes-sub2}
    \caption{Few shot classification on CUB}
    \end{subfigure}
    \caption{\textbf{Comparing performance with different number of similarity conditions on the test set.} For providing supervision with $n$ attributes (where $n$ may be less than the total number of attributes labelled in the dataset), the first $n$ of a fixed random order of attributes were chosen.
    }
    \label{fig:plot-nodes}
\end{figure*}

In Figure \ref{fig:plot-nodes} we plot the performance of the PAN-supervised and the PAN-unsupervised models with different numbers of similarity conditions. For supervision, we randomly shuffled the order of attributes and selected the first $n$.

In Figure \ref{fig:plot-nodes}(a), where we plot only the FITB accuracy of the models on the \emph{resampled} Polyvore Outfits split, we see that when there are few supervised attributes, the performance of the fully supervised model is poor. However, when the similarity conditions are allowed to be free (in the unsupervised model), the performance is higher. The supervised model starts performing better as the number of different attributes that are annotated increases. The increase in performance of the unsupervised model with increasing number of similarity conditions is much less pronounced, thus showcasing the role of supervision using attribute annotations in fashion compatibility prediction performance on Polyvore Outfits.

A different trend is seen in the case of few shot classification on CUB (in Figure \ref{fig:plot-nodes}(b)), where the unsupervised model has performance close to that of the fully supervised model. There is a general increase in performance with more similarity conditions in both the unsupervised and the supervised models, but the performance is high enough with a few similarity conditions. This indicates that good performance on the few shot classification task can be achieved with a few unsupervised similarity conditions, and supervision using attributes provides a small boost most of the time.  The relatively good performance on this task may be the result of the relative simplicity of the task, where few-shot classification can be thought of as trying to simply match attributes.  In comparison, in fashion compatibility, the relationship between attributes is far more complicated, as discussed in the introduction of our paper.  
Images with different attributes can be deemed highly similar (or more compatible), while some attribute combinations may indicate dissimilarity/incompatibility even if subsets of them would normally indicate similarity.
This requires a far more complex function to reason about attributes, which our results seem to indicate is difficult to capture without supervision. 

We also find that using supervised attributes often requires a critical mass, \ie, a variety of attributes are required to outperform the unsupervised model reliably. Our results on Polyvore Outfits demonstrate that these need not be dense annotations. Attributes on that dataset were automatically labeled after curating a set of visual concepts manually from common words that appear in the items' description and/or title, resulting in a very sparsely annotated dataset.

\section{Choosing number of similarity conditions for unsupervised models}

\begin{figure*}[t!]
    \centering
    \begin{subfigure}[t]{.39\textwidth}
    \includegraphics[width=\linewidth]{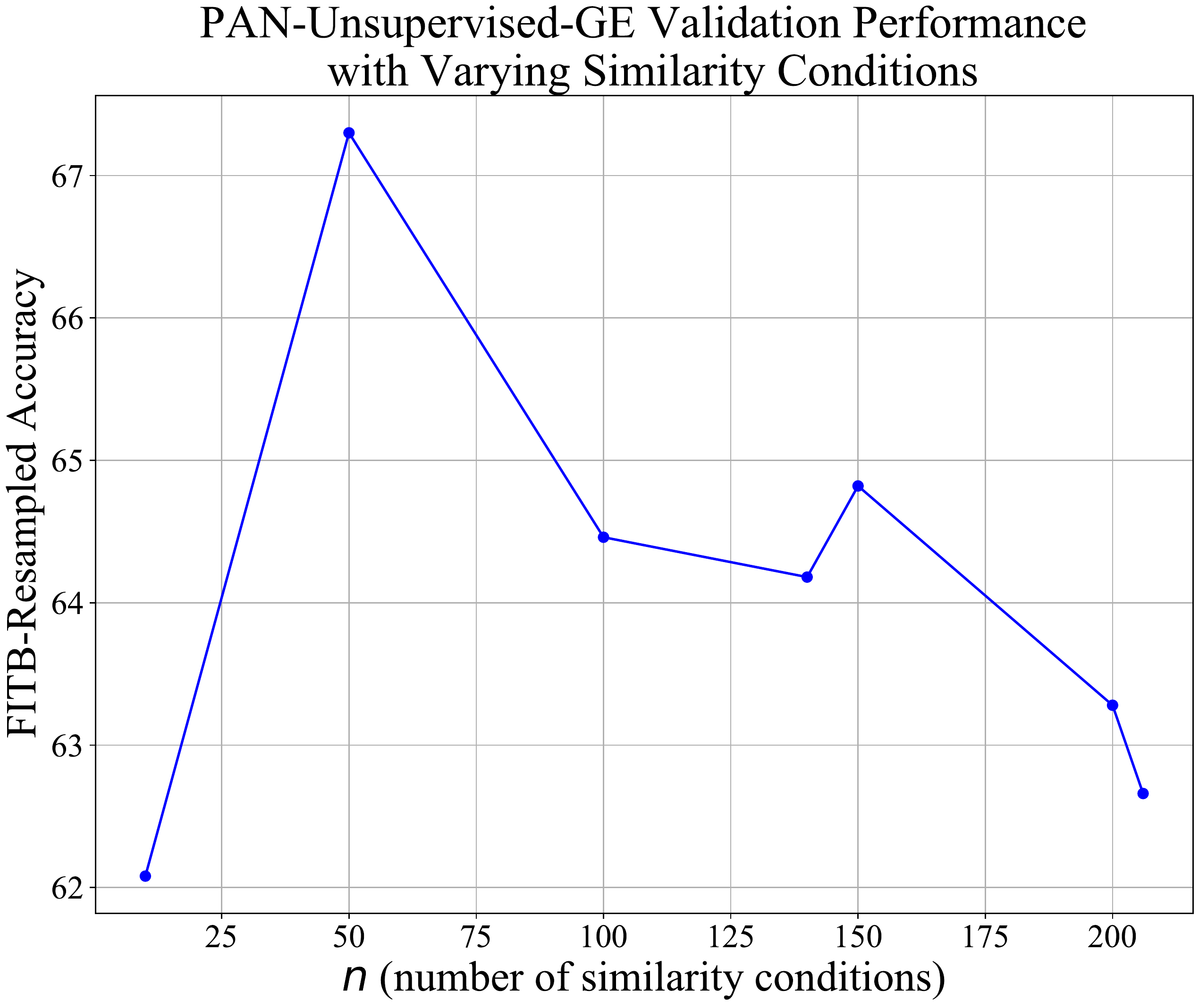}
    \label{fig:plot-nodes-val-sub1}
    \caption{FITB accuracy on Polyvore Outfits}
    \end{subfigure}
    \begin{subfigure}[t]{.58\textwidth}
    \includegraphics[width=\linewidth]{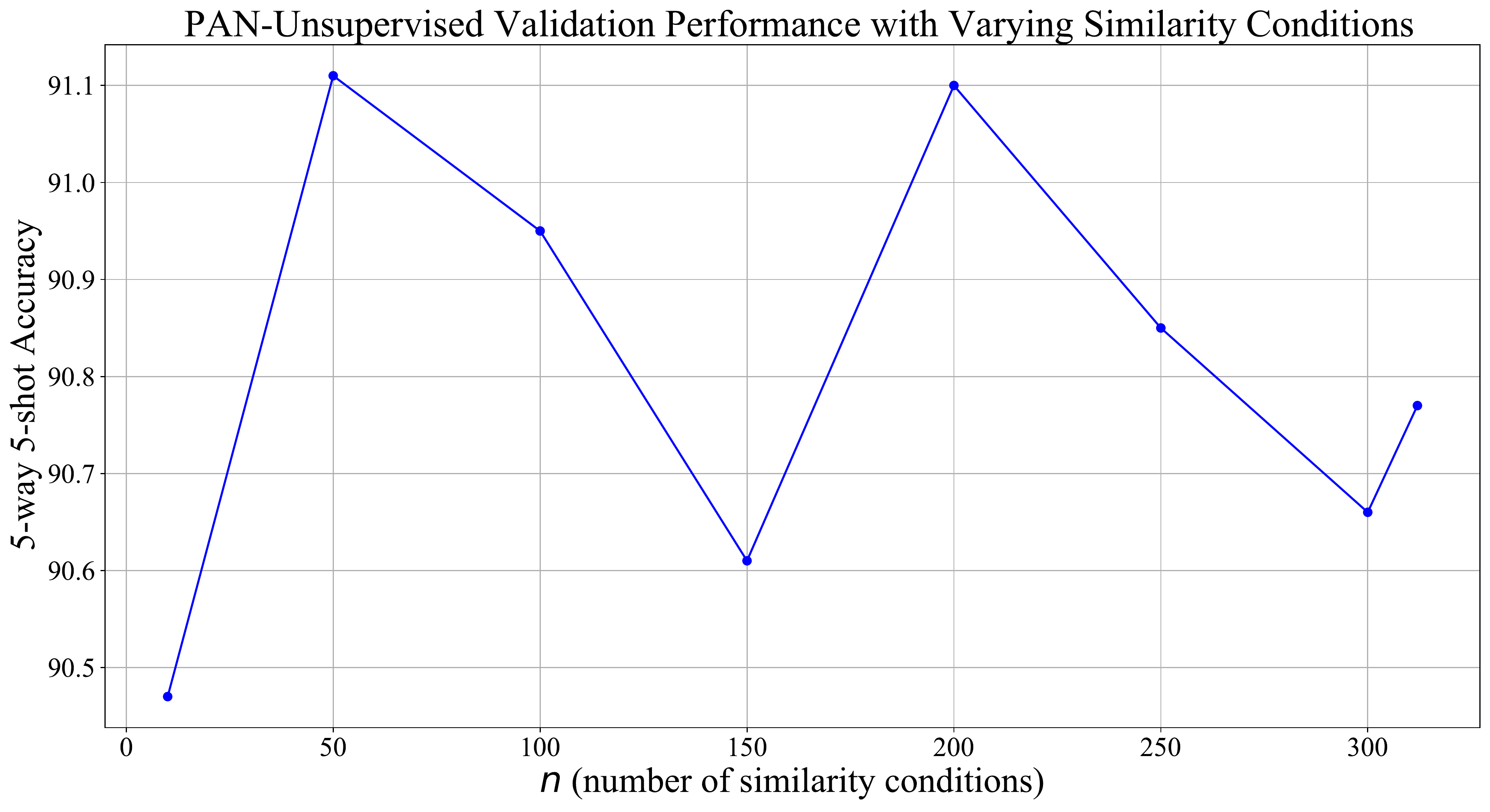}
    \label{fig:plot-nodes-val-sub2}
    \caption{Few shot classification on CUB}
    \end{subfigure}
    \caption{\textbf{Validation set performance} of the PAN models with unsupervised similarity conditions}
    \label{fig:plot-nodes-val}
\end{figure*}

Figures \ref{fig:plot-nodes-val}(a) \& \ref{fig:plot-nodes-val}(b), show the validation set performances of the unsupervised models in the two tasks---predicting compatibility on Polyvore Outfits and few shot classification on CUB. We chose the models with the best validation accuracy for comparison in Tables 1, 2 and 4 of the main paper. In particular, for both tasks, we found that the best performance was achieved at $50$ unsupervised similarity conditions ($M = 50$) and beyond that the model seems to overfit.

\section{Attribute Recognition Experiments} \label{sec:attr_recog}
We train the PAN models using supervision from the logical OR of attributes, and have seen that this helps in improving similarity prediction performance. Here, we inspect if the supervision results in meaningful predictions along different supervised similarity conditions. On pairs of images from the test split of the data, we compute the average precision (AP) of the attribute scores output by our model where the ground truth comes from the OR of the attribute labels. Missing attribute labels and their score predictions are excluded. The mean AP (mAP) is then computed by averaging the APs over the different attributes.

\begin{figure*}[t!]
    \centering
    \begin{subfigure}[t]{.35\textwidth}
    \includegraphics[width=\linewidth]{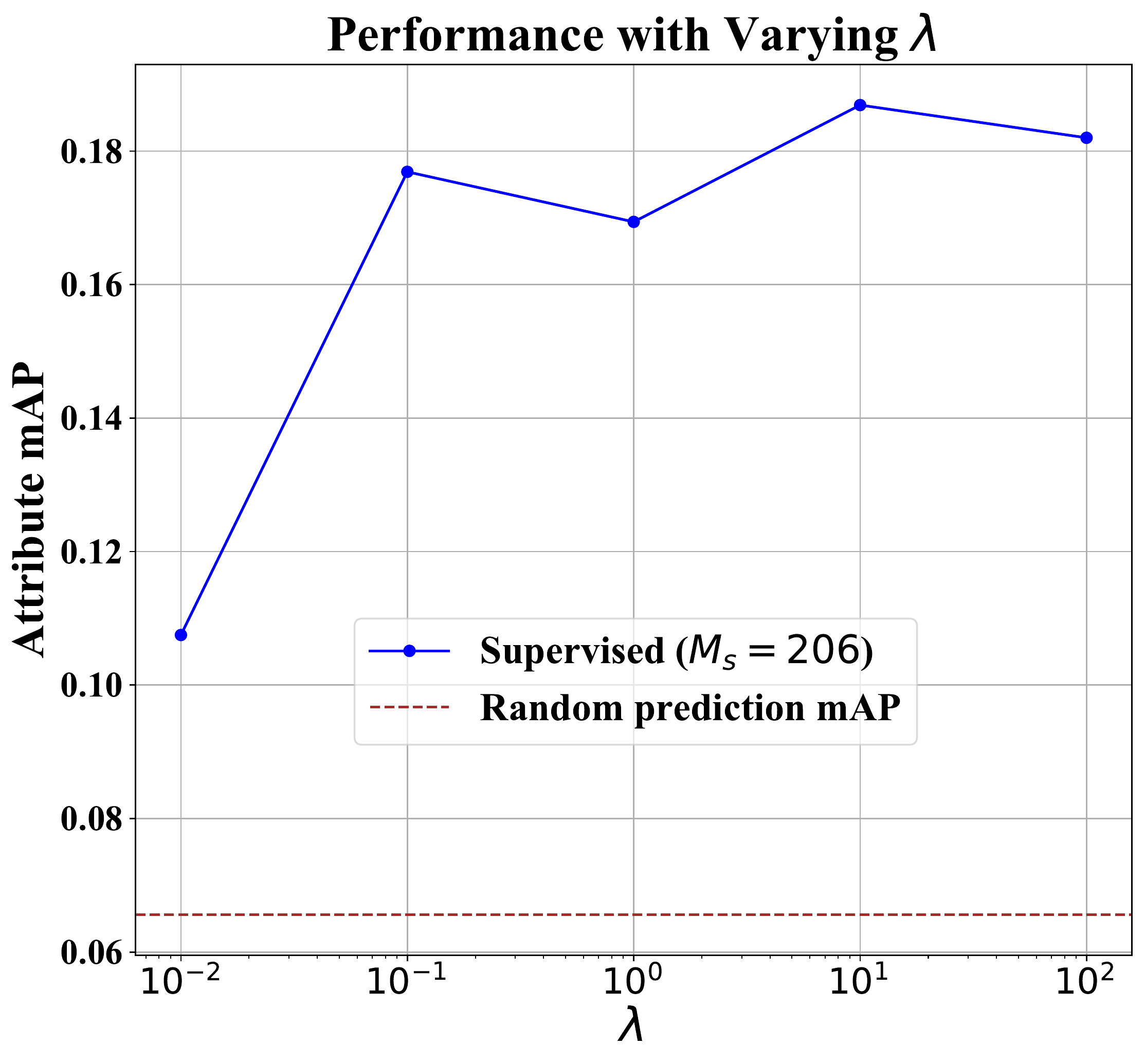}
    \label{fig:plot-lam-mAP-sub1}
    \caption{Attribute recognition on Polyvore Outfits.}
    \end{subfigure}
    \begin{subfigure}[t]{.61\textwidth}
    \includegraphics[width=\linewidth]{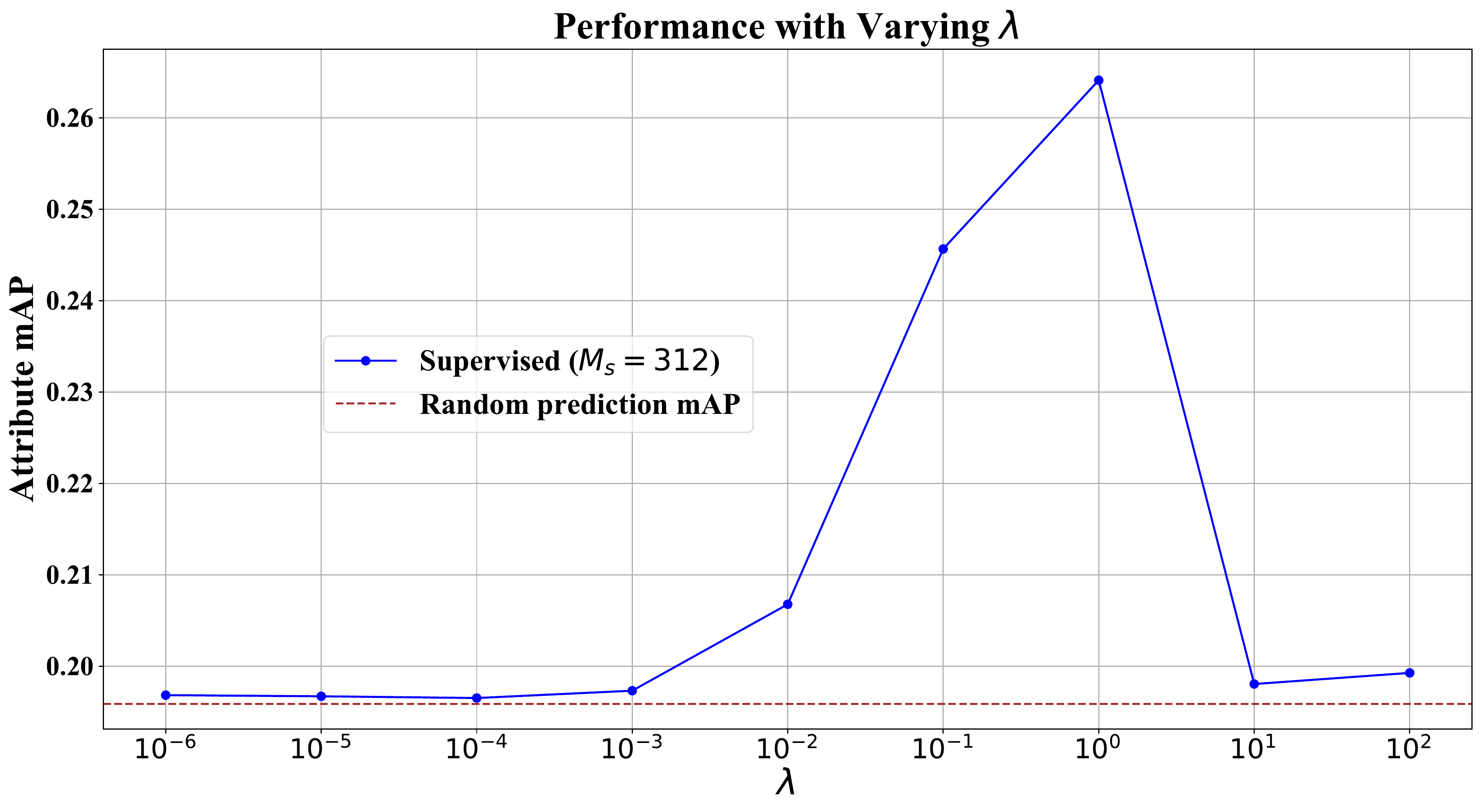}
    \label{fig:plot-lam-mAP-sub2}
    \caption{Attribute recognition on CUB}
    \end{subfigure}
    \caption{\textbf{Mean average precision (mAP) of attribute prediction.}}
    \label{fig:plot-lam-mAP}
\end{figure*}

In Figure \ref{fig:plot-lam-mAP}, we compare the mAP values as described above for our PAN-Supervised model using different values of the hyperparameter $\lambda$. In both Figures \ref{fig:plot-lam-mAP}(a) and \ref{fig:plot-lam-mAP}(b), the brown dotted horizontal line is the mean average precision of a set of scores generated uniformly at random from the range $[0, 1]$. From both figures, we see that increasing the weight of attribute supervision increases the mAP. 
Combining the results of Figure \ref{fig:plot-lam-mAP} with those of Figure 3 in the main paper, on the Polyvore Outfits dataset we see a positive correlation between the end-task performance and the mAP on attributes. $\lambda = 10$ results in both the best end-task performance as well as the best mAP.  This suggests that if we were to improve our model to better recognize joint attributes, we would improve compatibility predictions as well. On the CUB dataset however, we see a possible trade-off, where the attribute prediction performance is better for a higher lambda, but end-task performance is better for a lower value of lambda. This is in line with what we observe in Section \ref{subsec:nodes}, where we see a smaller role of attribute supervision in improving few shot classification performance on CUB as compared to its effect in fashion compatibility prediction on Polyvore Outfits.

\section{More questions} \label{sec:more_questions}
\noindent\textbf{How useful are relevance weights to model performance?} To find out, we experimented with an approach that predicts output similarity scores as PAN does and simply averages them to get the final similarity prediction between two images. On CUB, the 5-way 5-shot accuracy of this model was $85.25 \pm 0.28$ (over 3 runs with different random initializations) compared to $92.77 \pm 0.30$ with the relevance weighted sum. The FITB accuracy on the resampled split of PO, for it was $63.6$ compared to $69.7$ with the relevance weights. Thus simply averaging similarity predictions was found to perform poorer, highlighting the importance of using relevance weights.

\noindent\textbf{What is the reason behind using the logical functions that were used for attribute combination $f_a$?}
We defined $f_a : \{0,1\} \times \{0,1\} \rightarrow [0, 1]$ as some function that maps two input attribute labels to a pair-wise label for supervision. The definition lends quite some flexibility in choosing what $f_a$ can be, where possibilities include both binary (\ie outputs only take on values $0$ or $1$), or fractional outputs (in the entire range $[0, 1]$). $f_a$ can be either a non-parametric function, or could involve parameters (\eg it could be a linear combination or could be non-linear like a multi-layer perceptron). To narrow down this range of choices, for our experiments, we restrict $f_a$ to only binary values leaving the exploration of other choices to future work. Fractional outputs of the attribute combination $f_a$ could have benefits in certain scenarios, for instance, it could be used for ranking images by some form of attribute strength as done in ``relative attributes'' \cite{parikh2011relative}. 

Restricting $f_a$ to binary outputs results in a total of 16 possibilities. Out of the 16, ruling out functions that are non-commutative, there are 8 possible choices remaining. This set largely consists of common logical functions. We can further narrow this down to 4 choices, which are the ones we experimented with in Section 5 of the main paper. Of the 8, the functions left out were the constants (always 0 or always 1) and $NAND$ and $NOR$. The first two do not provide information regarding the inputs, and the latter two are simply negatives of $AND$ and $OR$ respectively.

As a form of sanity-check, we also experimented with randomizing the attribute labels to ensure that supervision indeed helps (even if it is sparse), rather than there being some form of a regularizing effect from random labels. On the resampled split of PO, the FITB accuracy of a model that was trained with such random labels was $58.0$, compared to $69.7$ for a supervised model with OR attribute combination, verifying the role of attribute supervision.

\noindent\textbf{Is it possible to train a model with both supervised and unsupervised similarity conditions?} An excellent question and certainly a possibility. However, in our experiments, we found that such a model could (ab)use its additional capacity to overfit to training data. On the CUB dataset, a ``hybrid'' model with 312 each of supervised and unsupervised similarity conditions could achieve a 5-way 5-shot accuracy of $91.38 \pm 0.24$. On the resampled split of PO, a model with 150 each supervised and unsupervised conditions had a FITB accuracy of $62.90$. Note that in this experiment, we had to restrict to fewer than 206 attributes because of GPU memory constraints, but even so, this model had more (a total of 300) similarity conditions than both the PAN-Supervised and PAN-Unsupervised models reported in Table 1 of the main paper. We leave to future work, a more in-depth analysis of this model and a possible way of effectively utilizing both unsupervised and supervised similarity conditions.

\begin{figure}[t]
    \centering
    \begin{subfigure}[t]{0.5\textwidth}
    \centering
    \includegraphics[width=\linewidth]{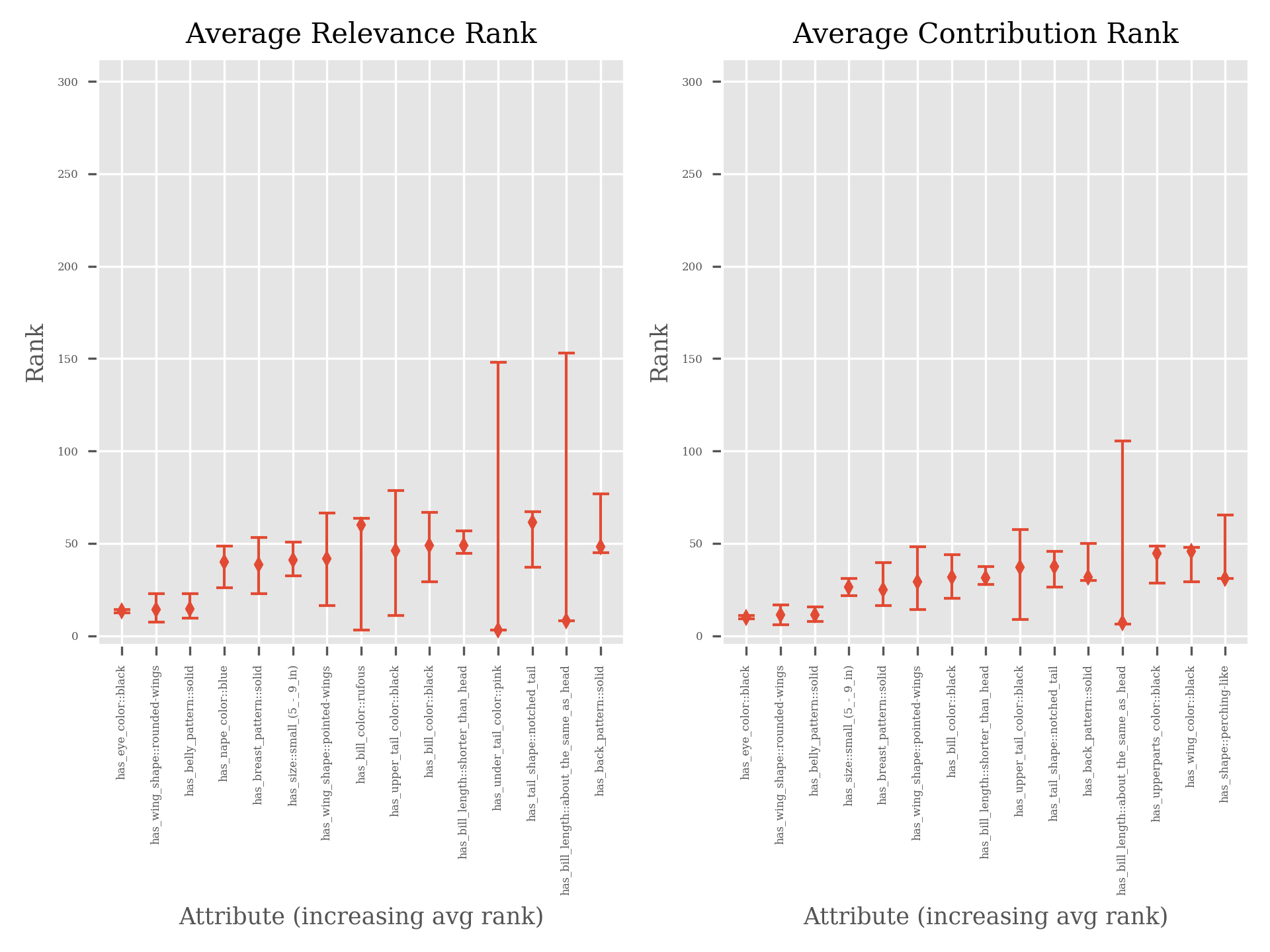}
    \caption{Attributes sorted according to increasing average rank}
    \end{subfigure}
    \begin{subfigure}[t]{0.5\textwidth}
    \centering
    \includegraphics[width=\linewidth]{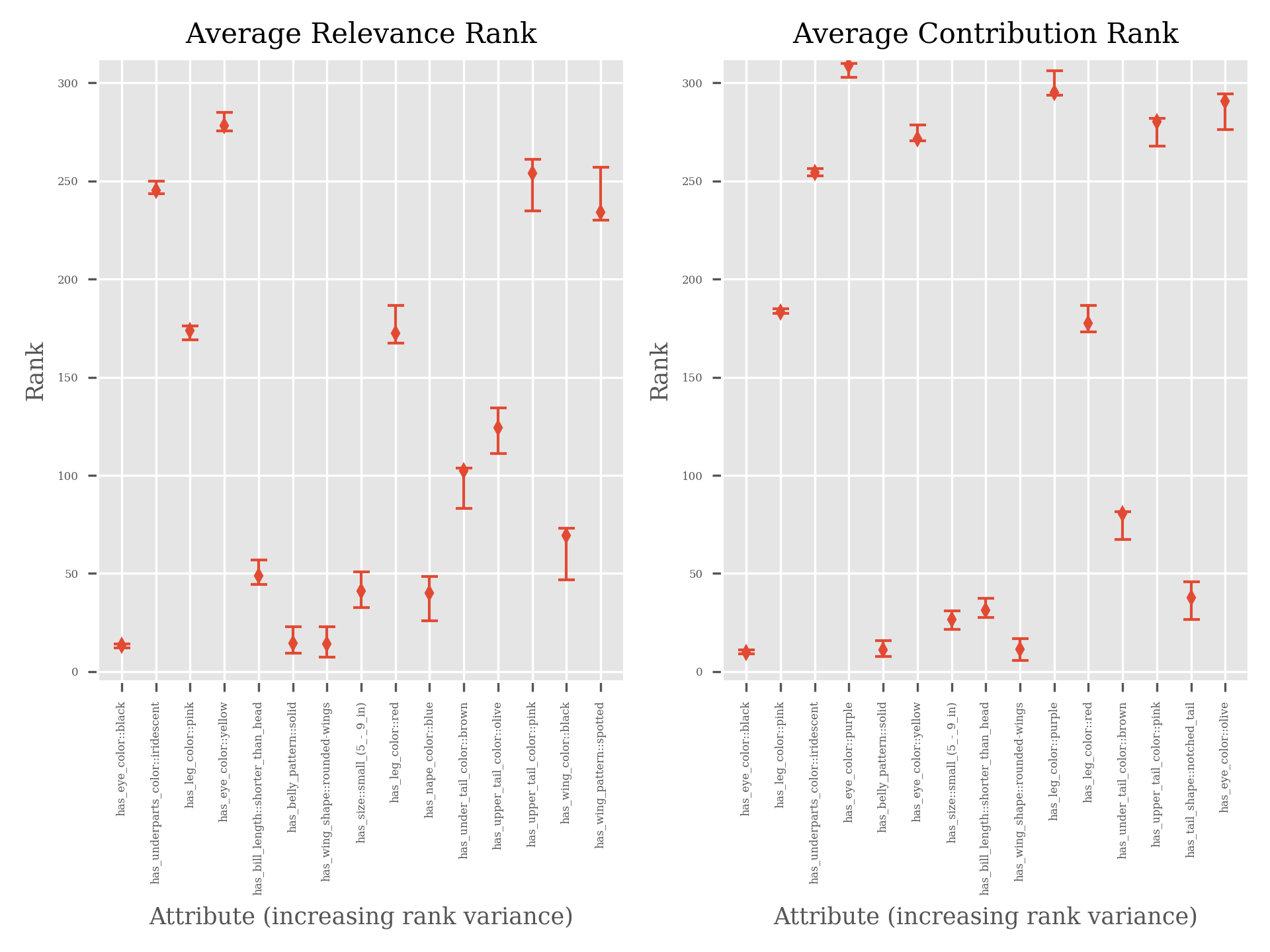}
    \caption{Attributes sorted according to increasing variance}
    \end{subfigure}
    \caption{Attributes' mean ranks and standard deviations by relevance score and contribution to the similarity score over multiple training runs. See Sec. \ref{sec:more_questions} for discussion. (Best viewed under zoom)}
    \label{fig:attr_ranks}
    \vspace{-5mm}
\end{figure}

\noindent\textbf{Do the attribute importance scores always reflect human intuition?}
On the CUB dataset, we found some variance across the attribute importance scores for the same examples in different training runs starting from the same initializations. Fig \ref{fig:attr_ranks} shows the variance in average ranks of different attributes across 3 different training runs of PAN-supervised. Ranks are computed based on both the relevance score (left) and the contribution to the similarity score (right) with the highest score being rank 1. The average ranks are computed by first computing for each run, the average rank of a given attribute in terms of its score from the PAN model across all pairs of images in the novel (or test) split of the dataset. This results in 3 average ranks (one per run). The average of these and the standard deviation (using error bars) are plotted in the figure. 

We see that after the first three attributes sorted by rank (Fig. \ref{fig:attr_ranks} (top)), the variance in the average rank increases by a lot. There are other attributes with smaller variance (Fig. \ref{fig:attr_ranks} (bottom)) but at a much higher rank (meaning they have low relevance or contribution). Thus, in the case of CUB, PAN has learnt a relevance predictor, that has fairly high variance dependent on initialization for most attributes. This behavior is likely a consequence of two factors: first, attribute labels were sparse and noisy and so PAN learned to treat some similarity conditions as latent or unsupervised. For some attributes, this variance in relevance score is also possibly a consequence of that attribute not being useful for determining similarity.  As a consequence, the model appeared to override these less important attributes to instead learn some general similarity metric.  This would manifest itself as that attribute having low recognition performance coupled with high contribution to the overall similarity score in some runs, but not others, resulting in high variance in an attribute's overall rank across initializations. Thus, we found PAN's attribute relevance predictions to be too noisy to be reliably consistent with human intuition, especially for the CUB dataset.

\end{document}